\documentclass[letterpaper, 10 pt, conference]{ieeeconf}  

\IEEEoverridecommandlockouts                              

\RequirePackage[2016/10/06]{latexrelease}
\usepackage{graphics} 
\usepackage{epsfig} 
\usepackage{mathptmx} 
\usepackage{times} 
\usepackage{amsmath} 
\usepackage{amssymb}  
\usepackage{array}
\usepackage{cite}
\usepackage{balance}
\usepackage{mathtools}

\title{\LARGE \bf
Context-Aware Multi-Task Learning for \\Traffic Scene Recognition in Autonomous Vehicles
}

\author{
{Younkwan Lee$^{1}$, Jihyo Jeon$^{1}$, Jongmin Yu$^{1,2}$ and Moongu Jeon$^{1}$} \\
\thanks{$^{1}$Machine Learning \& Vision Laboratory, Gwangju Institute of Science and Technology (GIST), Gwangju 61005, South Korea
        {\tt\small\{brightyoun, jihyo, mgjeon\}@gist.ac.kr}}%
\thanks{$^{2}$Curtin University, Kent St, Bently WA 6102, Australia
        {\tt\small jm.andrew. yu@gmail.com}}%
}

\begin{document}

\maketitle
\thispagestyle{empty}
\pagestyle{empty}

\begin{abstract}

Traffic scene recognition, which requires various visual classification tasks, is a critical ingredient in autonomous vehicles. However, most existing approaches treat each relevant task independently from one another, never considering the entire system as a whole. Because of this, they are limited to utilizing a task-specific set of features for all possible tasks of inference-time, which ignores the capability to leverage common task-invariant contextual knowledge for the task at hand. To address this problem, we propose an algorithm to jointly learn the task-specific and shared representations by adopting a multi-task learning network. Specifically, we present a lower bound for the mutual information constraint between shared feature embedding and input that is considered to be able to extract common contextual information across tasks while preserving essential information of each task jointly. The learned representations capture richer contextual information without additional task-specific network. Extensive experiments on the large-scale dataset HSD demonstrate the effectiveness and superiority of our network over state-of-the-art methods.

\end{abstract}

\section{INTRODUCTION}

    Traffic scene recognition from input images is one of the fundamental technologies for Automated Driving Systems (ADS) and Advanced Driver Assistance Systems (ADAS) applications, including object modeling \cite{azam2018object}, semantic segmentation \cite{cordts2016cityscapes}, object recognition \cite{cheng2019dense,dinh2018real,lee2019unconstrained}, localization and mapping \cite{ziegler2014video,bresson2017simultaneous,munir2018autonomous}, etc. It is perceived as an essential key-step towards understanding traffic scenes and serves to eliminate the gap between resulting performance and visual reasoning capability of human beings. Unlike the previous works which typically optimize a single objective function, it is not merely a general classification problem since what we are dealing with here is multi-objective optimization. For example, traffic scenes contain diverse scene attributes, such as road place, weather, road surface, and road environment. Also, the bottom-level of the road place consists of how they are annotated to temporal action labels inside its place. Thus, a prominent challenge in traffic scene recognition is to make a model that can operate with online frame-wise inference despite multiple tasks.

    \begin{figure}[thpb]
      \centering
        \includegraphics[width=3.2in]{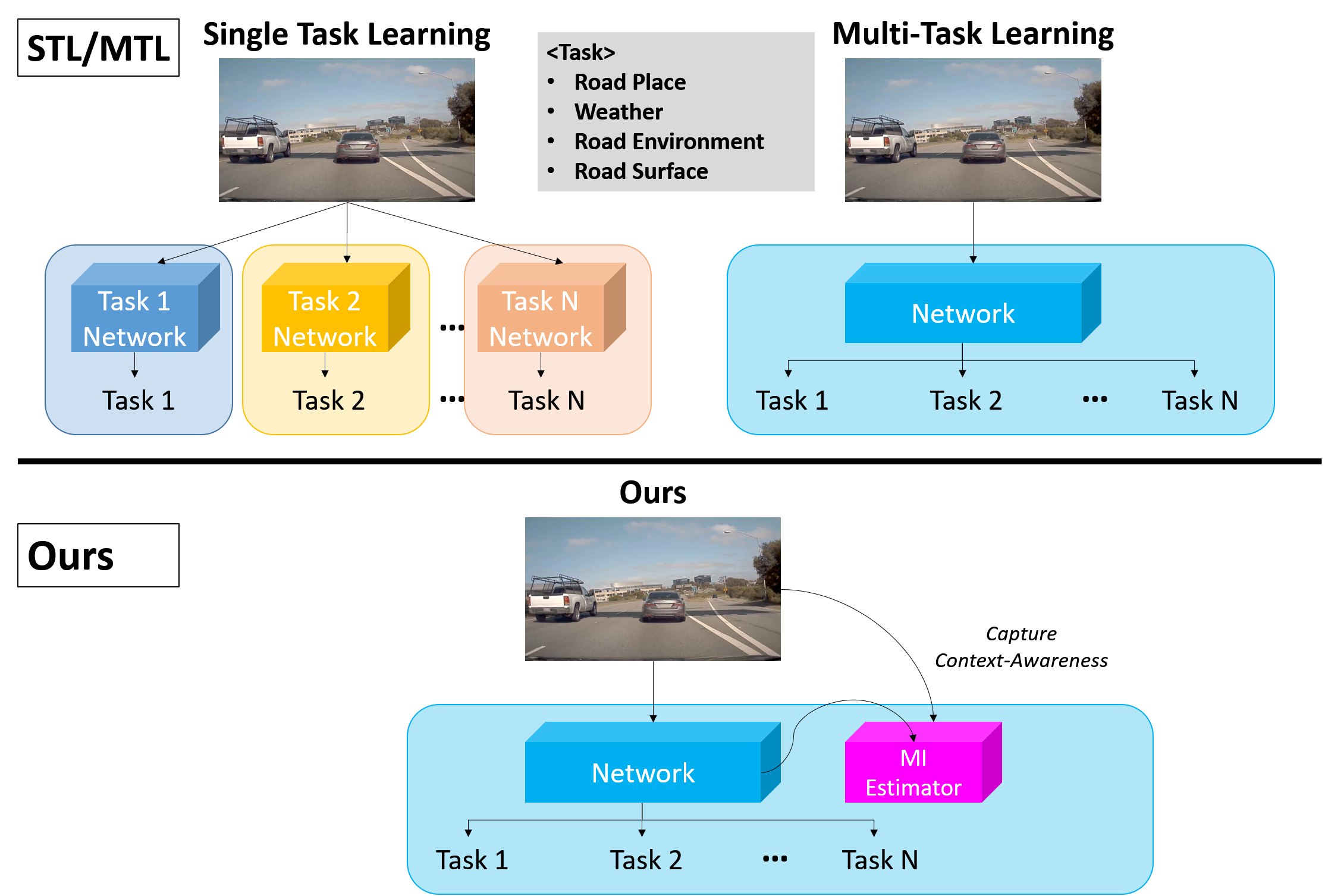}
      \caption{Comparison of the existing method and ours. The existing method consists of single-task learning (STL) and multi-task learning (MTL), these methods are not prone to be captured contextual information. Our method addresses this issue by employing the mutual information (MI) constraint and learn context-awareness ability.}
      \label{figurelabel}
    \end{figure}

    A na\"ive algorithm capable of performing multiple tasks simultaneously is to employ a suite of independent networks, one for each task. However, simply training a single network for a particular task, no matter how adequately optimized on each task, is far from satisfactory in terms of performance on every task. The reason is that such approach does not share any essential information that must be considered among tasks.

    A solution to this difficulty is to design a multi-task learning (MTL) network, where multiple tasks are solved simultaneously, assuming the same input. Recently, the emerged MTL studies \cite{Dvornik_2017_ICCV,teichmann2018multinet,chowdhuri2019multinet,chennupati2019multinet++} show benefits on visual scene recognition. These methods rely on the sharing of the encoder’s representation using auto-encoder-based Convolutional Neural Networks (CNNs) so that decoders corresponding to each task can be applied to independent targets. Although previous works on the MTL have made great efforts on enhancing the performance of multiple recognition tasks within limited parameters, there still remain gaps when in dynamic driving scenarios. The reason is that the sufficient contextual information from changing spatial and temporal relationships is not adequately exploited due to no more either a deeper and larger network or well-chosen hyper-parameters. 

 \begin{figure*}[thpb]
\begin{center}
   \includegraphics[width=7in]{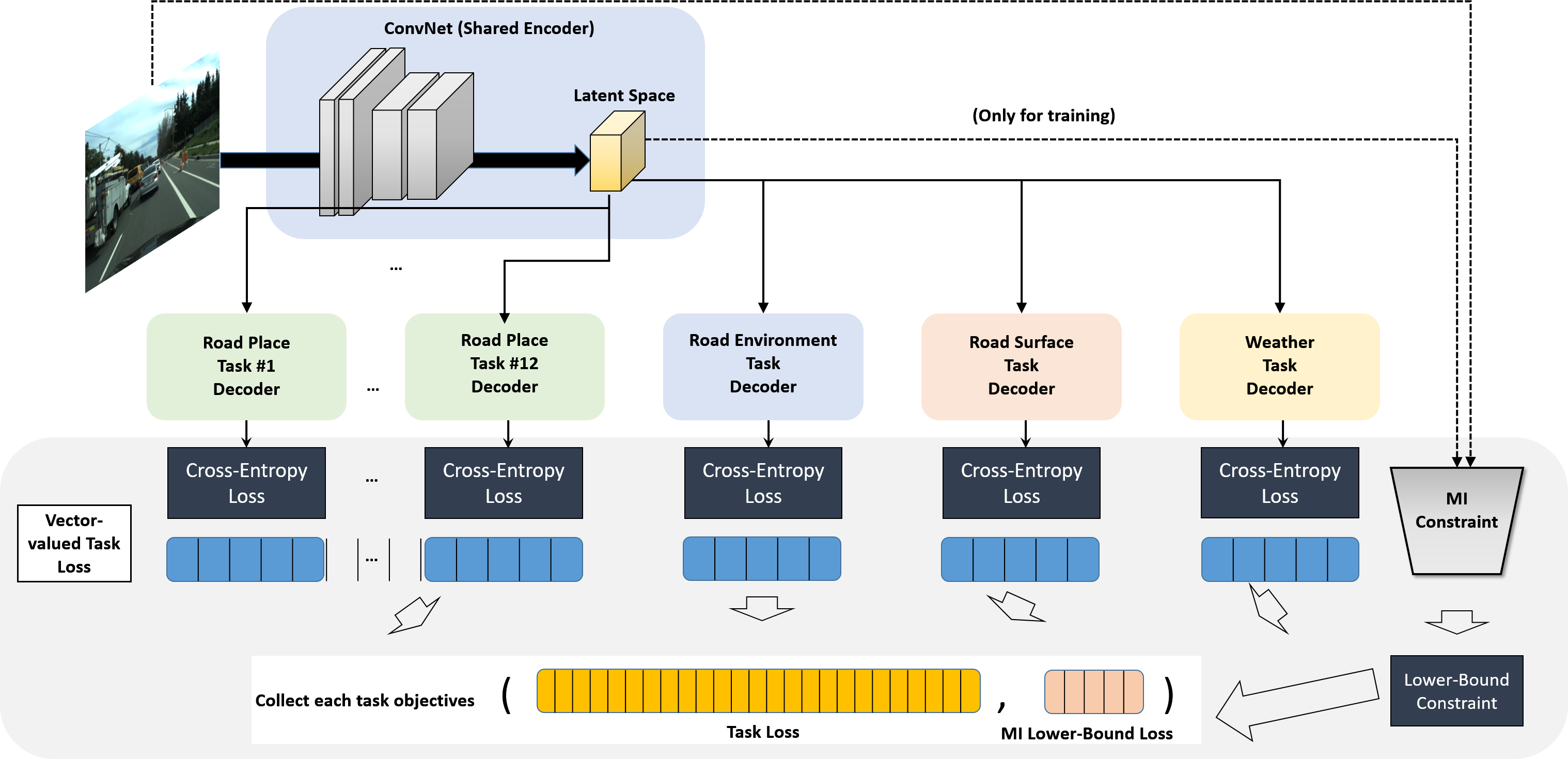}
\end{center}
   \caption{The pipeline of the proposed MTL traffic scene recognition system. The images are fed into the network; Shared encoder extracts latent space using MI constraint estimator; The task-specific decoders from the latent space utilize cross-entropy loss, respectively; Finally, the objective multi-task loss function is jointly trained by multi-task optimization.}
  \vspace{-0.5cm}
\label{fig:asdasd}
\end{figure*}

    To gain a fundamental understanding of efficient structure inside these visual multiple tasks, we believe that scene context illustrates the human's ability to understand the scene, even though the visual signal representing the scene is disturbed or insufficient. This assumption comes from the fact \cite{oliva2007role} that visual context is vital in the human visual system. Thus, without complementary measures, if the shared encoder learns to operate multiple tasks, its features important for task-specific factors might be wrongly aligned, leading to incomplete convergence with low saturation. Every task-specific decoders try to assign appropriate features directly through the shared encoder, but weights of the previous task might be shifted, whenever focusing on the current task. Unfortunately, repeating the above convergence, they are likely that an entire framework would show deterioration in performance termed as "\textit{catastrophic forgetting}" \cite{mccloskey1989catastrophic, ratcliff1990connectionist, Goodfellow14anempirical}.

    In this paper, we propose a context-aware MTL network, based on mutual information estimation, for traffic scene recognition in autonomous vehicles. As shown in Fig. 1, we start from a general MTL-based scene recognition setting, where an image classifier based on CNN framework (\textit{e.g.} ResNet \cite{he2016deep} and DenseNet \cite{huang2017densely}) is given and scene annotations are available from the traffic scene dataset \cite{narayanan2019dynamic}. The core idea of our method is to estimate mutual information constraint that calculates the lower-bound between input and task-specific latent space, leveraging shared encoder. Therefore, we refer to it as a new regularization loss term that enforces the MTL framework to be better context-aware features for easier recognition. This capture common contextual information across the entire feature representation including shared encoder and decoders while considering its unique properties of input image. Moreover, in spite of learning the way of decoders targeted on only own task, our approach would acquire good task-specific features in each decoder, maintaining high-quality context-aware features. We empirically show that the proposed approach is substantially effective for accurate scene recognition.

    We highlight our key contributions as follows: First, we propose a novel multi-task learning (MTL) framework for traffic scene recognition, where a shared encoder and decoders for each task are introduced to perform multi-objective optimization. Second, we introduce a novel mutual information-based regularization loss to promote the MTL network within no additional network parameters while preserving task-specific features and generating common context-aware features simultaneously. Finally, we demonstrate the effectiveness of our approach, which acquires the useful context-aware feature space and shows that the multi-task recognition performance outperforms other state-of-the-art methods, especially on the online inference.

\section{RELATED WORKS}

 \begin{figure*}[thpb]
\begin{center}
   \includegraphics[width=7in]{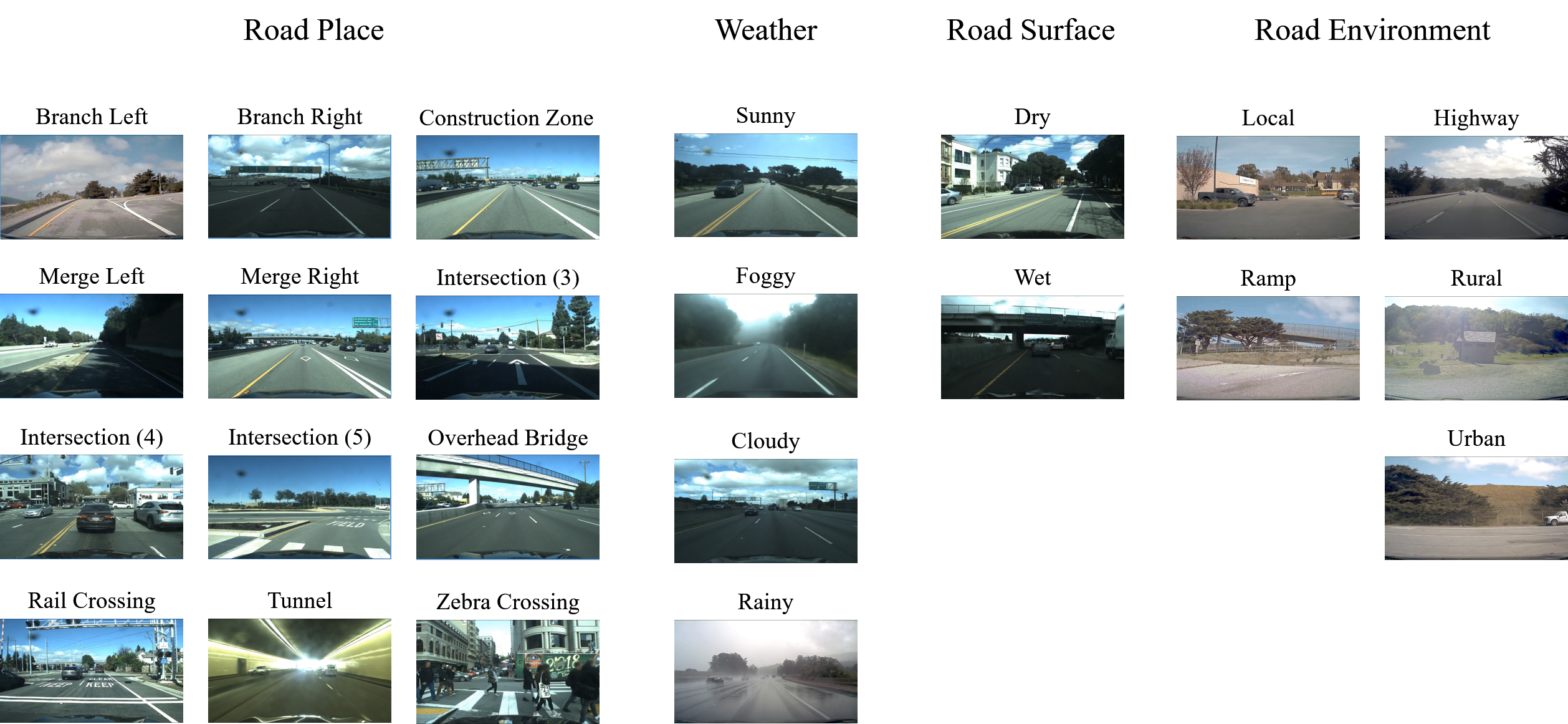}
\end{center}
\vspace{-0.5cm}
   \caption{Sample images of HSD dataset. Top 4 rows mean parent task. Road place, Weather, Road Surface, and Road Environment include 12, 4, 2, and 5 tasks, respectively.}
\label{fig:asdasd}
\end{figure*}

\subsection{Scene Recognition} 
    Scene recognition has been a topic of considerable research in the fields of the vehicle surrounding environment perception due to its important role in a wide variety of ADAS and ADS applications. Early works employed hand-crafted feature-based methods \cite{parodi1995feature,siagian2007rapid,song2009biologically,shroff2010moving,derpanis2012dynamic,feichtenhofer2016dynamic} to extract spatiotemporal information of the scene. However, most of the scene recognition systems based on hand-crafted features rely heavily on heuristic priors and low-level features, thus resulting in not robust to complex environments. Moreover, relatively insufficient traffic-scene datasets compared to general-scene datasets (\textit{e.g.} Maryland \cite{shroff2010moving}, YUPENN \cite{derpanis2012dynamic} and SUN \cite{xiao2010sun}) restricted the performance of the methods, especially in unconstrained settings.
    
    Recently, with increasing research interests in traffic scene understanding, some large-scale datasets \cite{sikiric2014image,jain2015car,xu2017end,ramanishka2018toward,narayanan2019dynamic} have been published in the past few years. Besides, several methods \cite{narayanan2019dynamic,ramanishka2018toward,wu2019deep,chen2019deep,sikiric2019traffic} focus on the deep neural architectures, inspired by the great success of image classification. \cite{sikiric2019traffic} developed a multi-label classification framework that is suitable for many mobile agents through short image descriptors. \cite{narayanan2019dynamic,chen2019deep} proposed large-scale datasets for the dynamic driving scene and designs multi-label architecture including event proposal network and resolution-adaptive mechanism so that showed better performance than the previous methods. Nevertheless, most of such methods have not been concentrated on explicitly extracting contextual information. This can lead to problems of catastrophic forgetting, where learned feature representations can gradually end up worse recognition performance.
    
    In this work, our proposed model adopts a context-aware learning approach. Although numerous methods are using the deeper and wider CNN module for better performance, our approach differs in the view of explicitly defining the degree of contextual information, which has not fully addressed in existing methods.

\subsection{Multi-Task Learning.} 
    Multi-task learning (MTL) addresses the problem of conducting parameter sharing across the tasks. For surveys of this field, we recommend reviews by \cite{ruder2017overview,zhang2018overview} on the basis of the proposed network structure that is most closely our work. As we add the tasks of numerous traffic situations for the human-like visual understanding of autonomous vehicles, the number of task-specific networks to optimize is proportional. Therefore, the importance of MTL multiple tasks at the same time is obviously growing because MTL reduces the computation needed for inference as well as convergence time for training. Existing deep learning-based MTL methods in computer vision \cite{misra2016cross,rudd2016moon,kokkinos2017ubernet,atapour2019veritatem,teichmann2018multinet,lee2019snider,Yogamani_2019_ICCV} demonstrated the efficiency, highlighting the extensibility for future developments and applications. A few works \cite{hong2018multimodal,yang2018end,chowdhuri2019multinet} considered the multi-modal inputs where multi-task network commonly handles multiple types of input. None of these methods attempts to explicitly define contextual information from the relationship between latent space and input. Our work focused on calculating neural mutual information for calculating contextual information. It is worth noting that, to our best knowledge, our work may be the first work to apply contextual information estimation for MTL.


 \begin{figure*}[thpb]
\begin{center}
   \includegraphics[width=7in]{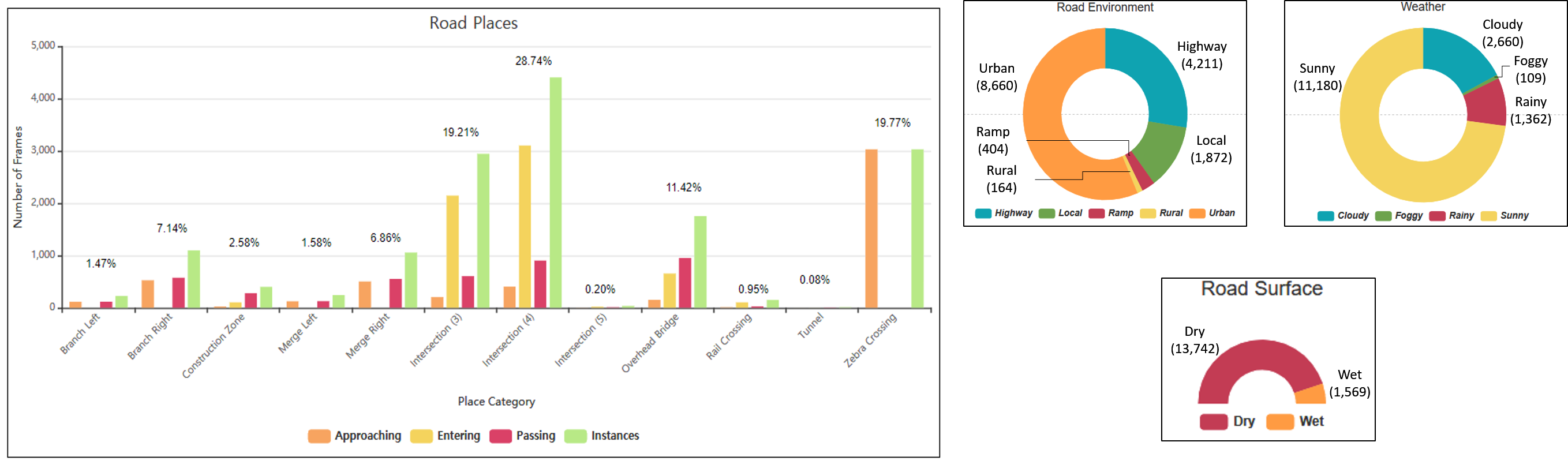}
\end{center}
\vspace{-0.5cm}
   \caption{The statistics of tasks in HSD dataset.}
\label{fig:asdasd}
\end{figure*}

\section{PROPOSED METHOD}
    In this section, we describe our approach to traffic scene recognition. First, we illustrate problem settings and the overall idea for the problem. Next, we present the proposed mutual information constraint. Lastly, we introduce a lower bound of each mutual information estimation across tasks and represent the final loss function for training the MTL network. Our MTL framework is shown in Fig 2.
    
\subsection{Formulation} 
    We focus on the problem of context-aware MTL using deep learning in traffic scene recognition. Then we have the multi-task training set $\mathbb{X} = \{x_1, x_2, ...,x_N\}$ of $N$ frames where the task spaces $\mathbb{Y} = \{y^1, y^2, ...y^T\}$ with the number of tasks $T$. We denote $x_i \sim p(x)$ as the random variables of input image $x_i$ on the $i-th$ frame. We consider each image sample to project into a latent representation $z \in \mathbb{Z}$ that is dynamically learned for a variety of tasks. To resolve this goal, we utilize a deep feature extraction function $E(\cdot;\theta_e)$ : $\mathbb{X} \rightarrow \mathbb{Z}$, with encoder network parameters $\theta_e$. Our encoders are learned and run with weight sharing for all tasks, while decoders are defined separately for each task. We acquire the representation $z$ from input image $x$ by sampling through conditional probability $p_{\theta_e}(z|x)$ with set of parameters $\theta_e$. 
    
    Regarding multi-task decoders, let $D^t$ denotes the task-specific network for neural hypothesis class per task that takes latent space $z$ as input. In this paper, we create decoders for a total of 23 tasks, including 12 in Road Place, 4 in Weather, 2 in Road Surface, and 5 in Road Environment. The decoder $D^t$ is parameterized as follows: $D(\cdot|\theta_{t})$. For $i = {1, ...,N}$, the $i$th image has MTL training labels: $x_{i} = \{(y^{(t)})\}^{T}_{t=1}$. Specifically, we employ multi-objective optimization based on Pareto optimality for eliminating the task-wise weights, which is a computationally inexpensive way. The multi-objective optimization for MTL using a vector-valued cross-entropy loss $\hat{L}$ is
    
     \begin{equation}
    	\begin{split}
    	    & \underset{\theta_e, \theta_1,...,\theta_T}{\min}L(\theta_e,\theta_1,...,\theta_T) = \frac{1}{N}\sum_{i=1}^{N}\frac{1}{T}\sum_{t=1}^{T}\hat{{L}}^t(D((E(x_i;\theta_e);\theta_j))).
    	\end{split}
     \end{equation}
     
     Although this objective is intuitively appealing, it often reveals the performance of MTL is not better than single-task learning on the multi-task dataset. To handle the issue, we propose to regularization for capturing contextual information in the region of latent representations, thus assisting MTL optimization.

\subsection{Mutual Information Constraint} 
    To create the context-aware latent representation, we adopt an information-theoretically constraint using mutual information. As previously discussed, the maintenance of contextual information for each task is essential. We encourage our encoder to extract context-aware features from the input image. Hence, the resulting latent representations of encoder would contain abundant information of input while preserving task-specific information. 
    We now propose a useful term that retains task-specific information and captures contextual information of input image. Generally $z_t \sim p_{t,\theta_e}$ defines as the random variables of representations from $t$-th task. Let rewrite the MTL objective formulation (1) limiting of mutual information $I(z_t, x)$ value for all of the task:
     \begin{equation}
    \begin{split}
    	 \underset{\theta_e, \theta_{1},...,\theta_{T}}{\min}&L(\theta_{e},\theta_{1},...,\theta_{T}) = \frac{1}{N}\sum_{i=1}^{N}\frac{1}{T}\sum_{t=1}^{T}\hat{L}^t(D((E(x_{i};\theta_{e});\theta_{t})))\\
    	 & s.t. \quad I(z_{t}, x) > \epsilon_{m}, \forall{t} \in \{1,...,T\}.
    \end{split}
     \end{equation}
     
     where the hyper-parameter $\epsilon_{m}$ controls the amount of mutual information. The equation (2) can be equivalently expressed with Lagrangian multipliers $\lambda_l$ as follows:
     \begin{equation}
    	\begin{split}
    	    & \underset{\theta_e, \theta_1,...,\theta_T}{\min}L(\theta_e,\theta_1,...,\theta_T) 
    	    \\& = \frac{1}{N}\sum_{i=1}^{N}\frac{1}{T}\sum_{t=1}^{T}\hat{L}^t(D((E(x_i;\theta_e);\theta_t))) - \lambda_l\sum_{t=1}^{T} I(z_t, x).
    	\end{split}
     \end{equation}
     
     However, objective (3) is not tractable in practice. Hence, we give a tractable approximation by using the lower bound of mutual information.

\subsection{Lower Bound of Constraint}
    Although task-specific patterns are ignored by equation (1), forcing the similarity in the marginal distribution does not directly affect the capture of valid information in each task. In particular, traffic scenes require task-specific information to be maintained, as the geometric and semantic context-aware representation of the image. Therefore, maintaining contextual information that is mapped by input images can contribute to the abundant preservation of the essential characteristics of the image. To solve this issue, we limit the minimum value of MI $I(z_t, X_i)$ for all tasks and this maintains the representation specified by all tasks.  
    
     Mutual information is lower bounded by Noise Contrastive Estimation \cite{oord2018representation}:
     \begin{equation}
    	\begin{split}
        	I(z_t, x) &\geq \hat{I}^{(NCE)}(z_t, x)
        	\\&:= \mathbb{E}_{p_{(x)}}[D(E(x;\theta_e);\theta_t)- \mathbb{E}_{\tilde{p}_{(x)}} [log\sum_{x^{'}}e^{D(E(x),x^{'})}]],
    	\end{split}
     \end{equation}
     where $x^{'}$ is sampled from the distribution $\tilde{p}(x)$ = $p(x)$. By providing stable approximation results \cite{hjelm2018learning}, mutual information can be maximized by replacing maximization in Jensen-Shannon divergence (JSD):
     \begin{equation}
    	\begin{split}
        	\hat{I}^{JSD}(z_t, x) := & \mathbb{E}_{p_{(x)}}[-sp(-D(E(x),x))] -
        	\\&\mathbb{E}_{p_{(x)} \times p_{(x^{'})}}[sp(D(E(x), x^{'}))],
    	\end{split}
     \end{equation}
     where $sp(\cdot)$ is a softplus function. As discussed in \cite{hjelm2018learning}, the decoder $D$ can share backbone encoder $E$ so that maximizing equation. (5) will maximize the MI $I(z_t, x)$. Finally equation. (3) can be rewritten as follows:
    \begin{equation}
    \begin{split}
    	    & \underset{\theta_e, \theta_1,...,\theta_T} \min L({\theta_e},{\theta_1},...,{\theta_T}) \\
    	    & = \frac{1}{N}\sum_{i=1}^{N}\frac{1}{T}\sum_{t=1}^{T}\hat{L}^t(D((E(x_i;{\theta_e)};{\theta_t}))) + {\lambda_l} \sum_{t=1}^{T}\hat{I}^{JSD}(z_t, x).
    \end{split}
    \end{equation}

 \begin{figure}[thpb]
\begin{center}
   \includegraphics[width=3.2in]{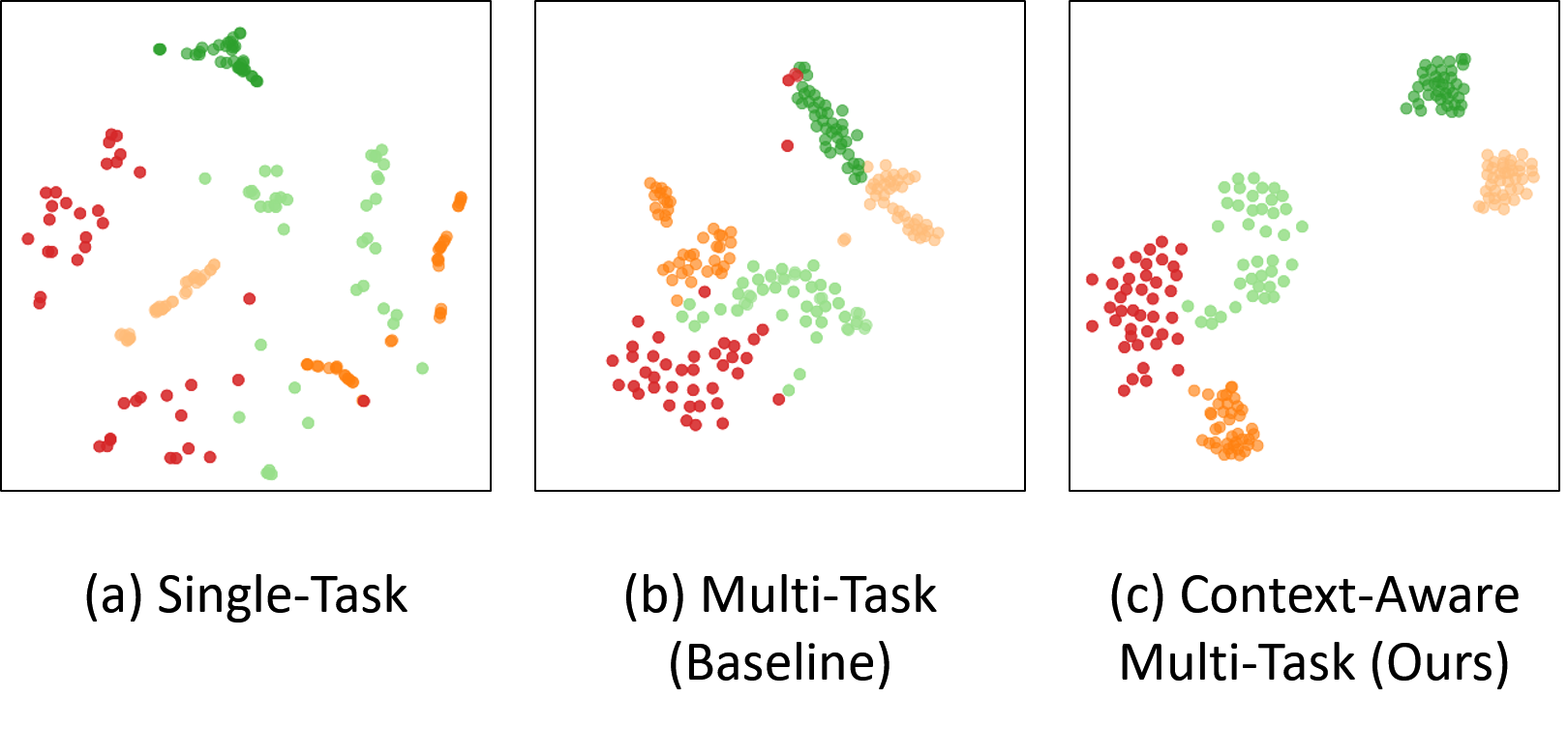}
\end{center}
\vspace{-0.7cm}
   \caption{Visualization for the embedding latent space learned by the single-task network, multi-task network (baseline) and context-aware multi-task network (Ours) on dataset HSD by t-SNE. Each point is corresponding to an input of a traffic scene. The different colors denote the different 'Road Environment' 5 categories.}
\label{fig:asdasd}
\end{figure}

 \begin{figure}[thpb]
\begin{center}
   \includegraphics[width=3.2in]{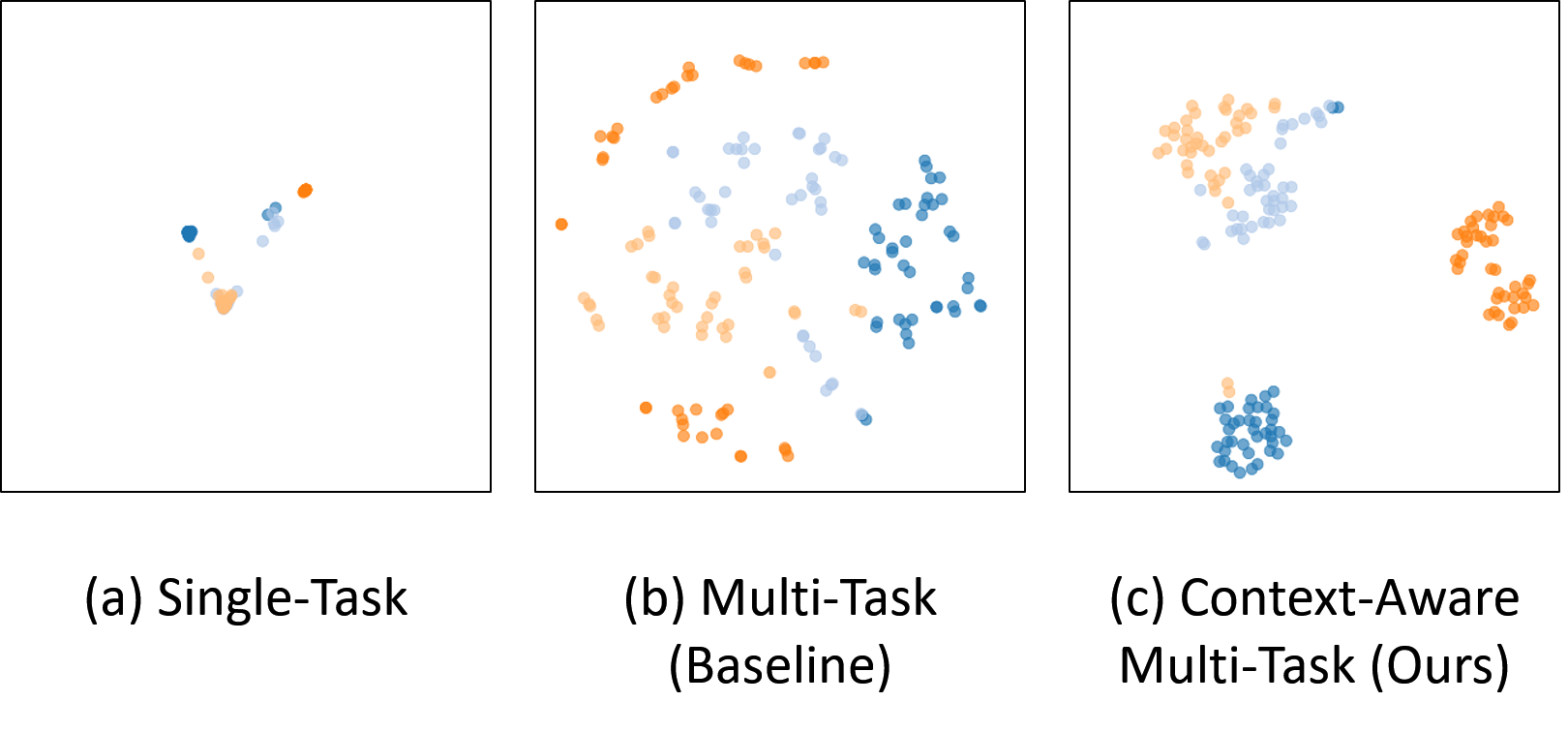}
\end{center}
\vspace{-0.7cm}
   \caption{Visualization for the embedding latent space learned by the single-task network, multi-task network (baseline) and context-aware multi-task network (Ours) on dataset HSD by t-SNE. Each point is corresponding to an input of a traffic scene. The different colors denote the different 'Weather' 4 categories.}
\label{fig:asdasd}
\end{figure}

 \begin{figure}[thpb]
\begin{center}
\vspace{-0.8cm}
   \includegraphics[width=3.2in]{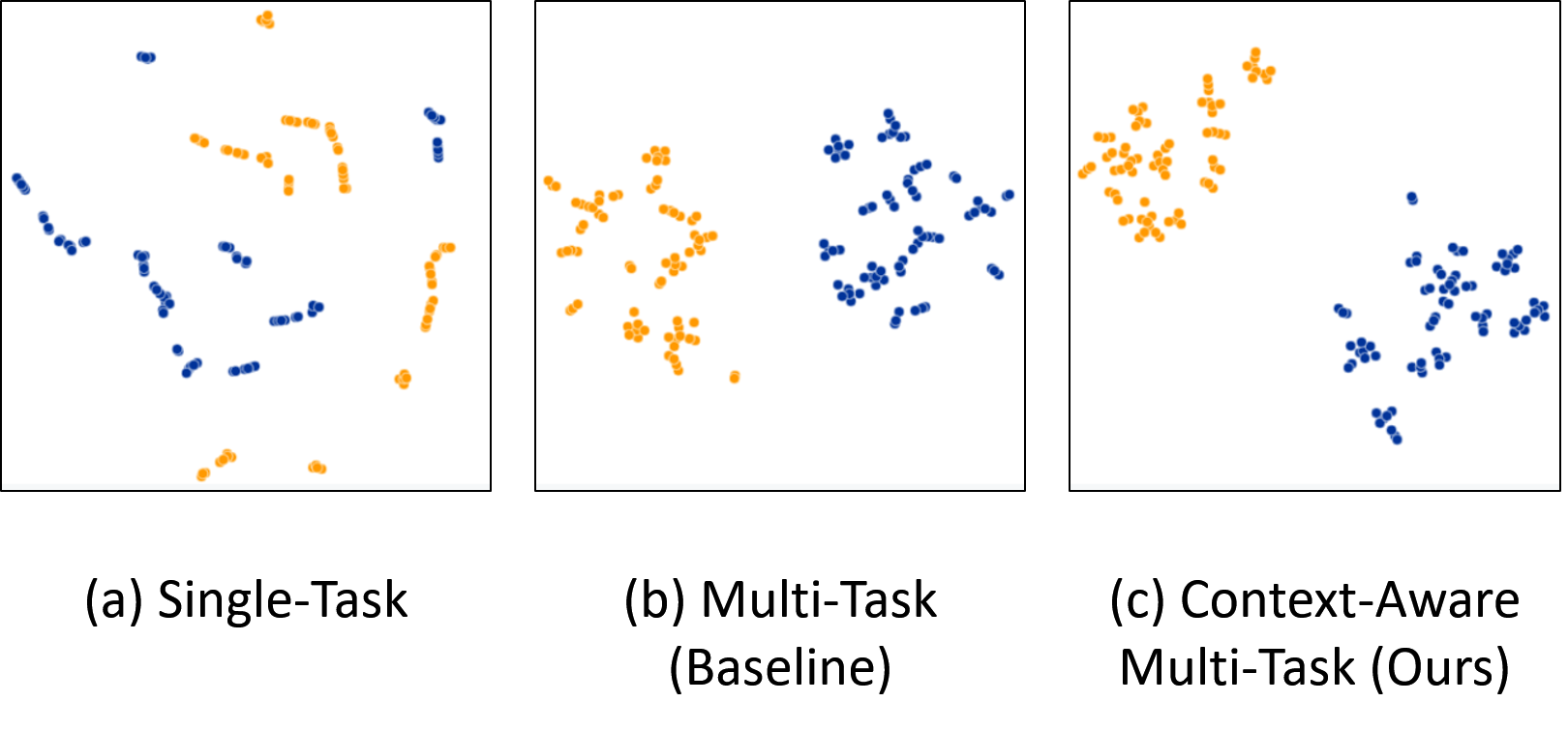}
\end{center}
   \caption{Visualization for the embedding latent space learned by the single-task network, multi-task network (baseline) and context-aware multi-task network (Ours) on dataset HSD by t-SNE. Each point is corresponding to an input of a traffic scene. The different colors denote the different 'Road Surfaces' 2 categories.}
\label{fig:asdasd}
\end{figure}

\section{EXPERIMENTAL RESULTS}

\subsection{Setup}
     All the reported implementations are based on the PyTorch frameworks, and our method has done on one NVIDIA TITAN X GPU and one Intel Core i7-6700K CPU. In all the experiments, we use a gradient clipping trick and the Adam optimizer \cite{kingma2014adam} with $\beta_1$ = 0.9, $\beta_2$ = 0.98, and weight decay to 0.0005. The proposed share encoder $E$ is implemented as a standard ResNet-50 architecture following the setting \cite{narayanan2019dynamic}. For the task decoders, we use a 2-layer CNN with global average pooling. In this paper, we deal with 23 tasks from HSD and optimize with only one objective function equation (6). The last convolutional layer is pooled to a size of class categories, respectively. All models adopt HSD dataset as the training set and are trained for the first 10 epochs with a learning rate of $0.0001$, and then for the remaining epochs at the learning rate of $0.00001$. Leaky-ReLU \cite{maas2013rectifier} and Batch normalization \cite{ioffe2015batch} are used in all layers of our networks with the mini-batch size set to 16 samples.

 \begin{figure*}[thpb]
\begin{center}
   \includegraphics[width=7in]{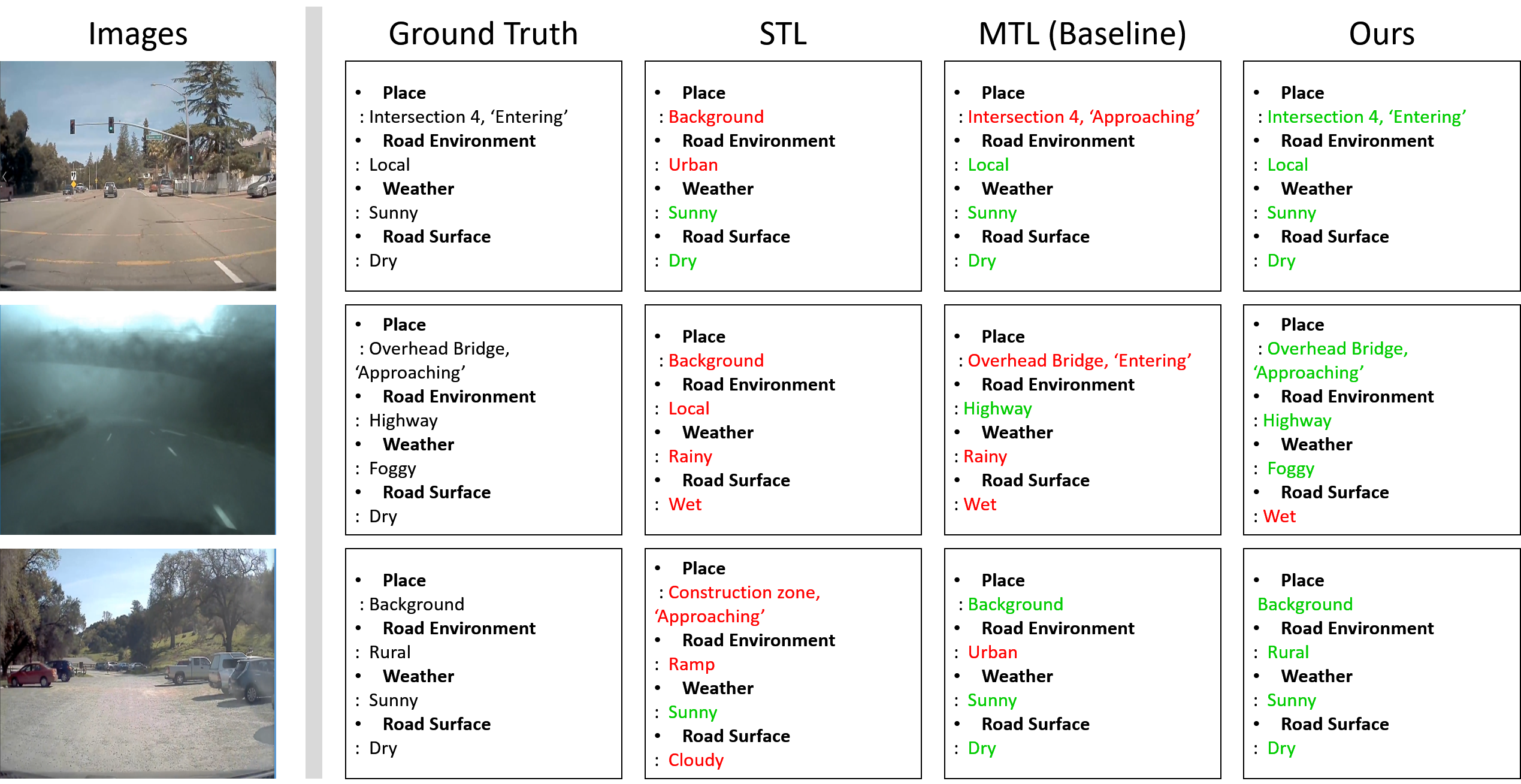}
\end{center}
\vspace{-0.5cm}
   \caption{Traffic scene images and the corresponding multi-task ground-truth labels. The evaluation results of single-task learning (STL) network, multi-task learning (MTL) network and our method. STL \cite{he2016deep} is the class-wise framework. In the results, the correctly predicted labels are denoted in green and the false predictions are denoted in red.}
\label{fig:asdasd}
\end{figure*}

\subsection{Dataset}
    The Honda Scenes dataset (HSD) \cite{narayanan2019dynamic} is a large scale annotated dataset and created to enable practical scene classification. The dataset contains 80 hours of diverse high quality driving video samples collected in the San Francisco Bay, USA. The dataset contains temporal annotations for road places, weather, road environments, and road surface conditions. 
    
    \textbf{Road Places} The dataset contains 12 classes of road places - branch with gore on left, branch with gore on right, construction zone, merge with gore on left, merge with gore on right, 3-way intersection, 4-way intersection, 5-way intersection, overhead bridge, railway crossing, tunnel, and zebra crossing. Most classes have 3 temporal sub-classes, including approaching, entering, and passing. The merge and branch classes have approaching and passing sub-classes, while the zebra-crossing class has just approaching.
    
    \textbf{Road Environments and Weather} The dataset spans 4 classes of road environments - rural, urban, highway and ramp and 4 weather conditions - rainy, sunny, cloudy, and foggy.
    
    \textbf{Road Surfaces} The dataset contains 2 classes of road surfaces - dry and wet.
    
    The dataset contains a total of about 20,000 instances spanned over these 12 classes. Moreover, the collected video images with camera are converted to a 1280 X 720 at 25 fps. The figure 3 outlines key statistics in the dataset.

 \begin{figure*}[thpb]
\begin{center}
   \includegraphics[width=7in]{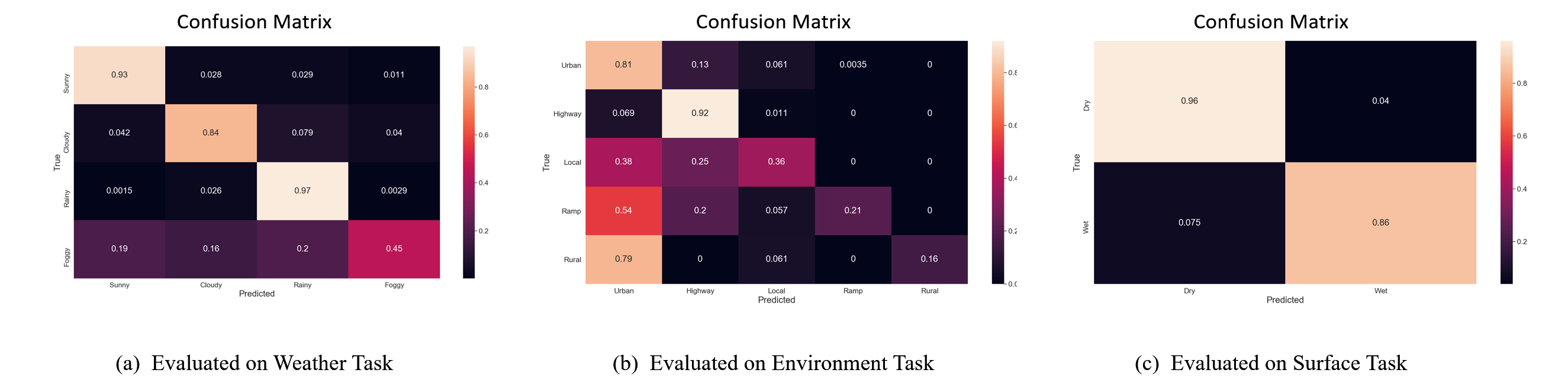}
\end{center}
\vspace{-0.7cm}
   \caption{Confusion matrices to visualize of the performance on each task. Lighter cells are higher values and the darker cells are lower values.}
\label{fig:asdasd}
\end{figure*}

\begin{table}[]
\caption{Ablation study on the effectiveness of our method. Type A and B represent the single-task network and multi-task network (Baseline), respectively. 4 Classes (Road place, Weather, Surface and Environment) are the result of mean F-score. For fair comparisons, the \textit{Tunnel}, \textit{Zebra Crossing}, \textit{Cloudy}, and \textit{Rural} classes, which were not used in previous studies, were excluded from the calculations.}
\label{tab:my-table}
\begin{tabular}{l|llll}
\hline
Type              & Road Place     & Weather        & Road Surface   & Road Environment \\ \hline
A                 & 0.285          & 0.910          & 0.950          & 0.560            \\
B                 & 0.277          & 0.908          & 0.936          & 0.580            \\ \hline
\textbf{C (ours)} & \textbf{0.303} & \textbf{0.916} & \textbf{0.950} & \textbf{0.601}   \\ \hline
\end{tabular}
\end{table}

\subsection{Ablation Studies}
     As our method can properly be extended to deal with multi-task, we compare our proposed method with the baseline traffic scene recognition researches to prove the effectiveness. Results on HSD are reported for the following three types of models where each approach is gradually added: a) Single-Task Network; b) Multi-Task Network (Baseline); c) adding MI constraint to (b) (Ours).
     
     We present the multi-task traffic scene recognition accuracy for each type on HSD in Table 1, and the evaluation results are shown in Figure 8. From Table 1, we can identify that expanding simply MTL network does not improve recognition performance. As shown in Figure 8., 'Place' category is difficult to recognize due to unbalanced dataset distribution. Therefore, when mutual information constraint is added to the proposed MTL, recognition results can be better (Figure 8. (last column)) and we observe some enhancements on traffic scene recognition performance (Table 1. e)). This confirms that applying MI constraint at the same time is more helpful to capture contextual information. Accordingly, the results in Figure 8 are the most accurate of all results.

\begin{table*}[]
\caption{F-Scores for each category of HSD. For each subcategory, its percentage in the entire dataset is also displayed under its name. Single-Task is based on ResNet-50 \cite{he2016deep} and Multi-Task is our baseline method. BL-\textit{Branch Left}, BR-\textit{Branch Right}, CZ-\textit{Construction Zone}, ML-\textit{Merge Left}, MR-\textit{Merge Right}, I-\textit{Intersection}, OB-\textit{Overhead Bridge}, RC-\textit{Rail Crossing}, T-\textit{Tunnel}, ZC-\textit{Zebra Crossing}.}
\vspace{-0.2cm}
\label{tab:my-table}
\resizebox{\textwidth}{!}{%
\begin{tabular}{l|clclcllclclcllcccccccccccccccc|}
\hline
 &
  \multicolumn{30}{c|}{\textbf{Road Place}} \\ \cline{2-31} 
\multicolumn{1}{c|}{} &
  \multicolumn{2}{c|}{BL} &
  \multicolumn{2}{c|}{BR} &
  \multicolumn{3}{c|}{CZ} &
  \multicolumn{2}{c|}{ML} &
  \multicolumn{2}{c|}{MR} &
  \multicolumn{3}{c|}{I3} &
  \multicolumn{3}{c|}{I4} &
  \multicolumn{3}{c|}{I5} &
  \multicolumn{3}{c|}{OB} &
  \multicolumn{3}{c|}{RC} &
  \multicolumn{3}{c|}{T} &
  ZC \\ \cline{2-31} 
\multicolumn{1}{c|}{} &
  A &
  \multicolumn{1}{c|}{P} &
  A &
  \multicolumn{1}{c|}{P} &
  A &
  \multicolumn{1}{c}{E} &
  \multicolumn{1}{c|}{P} &
  A &
  \multicolumn{1}{c|}{P} &
  A &
  \multicolumn{1}{c|}{P} &
  A &
  \multicolumn{1}{c}{E} &
  \multicolumn{1}{c|}{P} &
  A &
  E &
  \multicolumn{1}{c|}{P} &
  A &
  E &
  \multicolumn{1}{c|}{P} &
  A &
  E &
  \multicolumn{1}{c|}{P} &
  A &
  E &
  \multicolumn{1}{c|}{P} &
  A &
  E &
  \multicolumn{1}{c|}{P} &
  A \\ \hline
Bi-LSTM \cite{yue2015beyond} &
  \multicolumn{1}{l}{\textbf{0.36}} &
  \textbf{0.22} &
  \multicolumn{1}{l}{0.28} &
  \textbf{0.28} &
  \multicolumn{1}{l}{\textbf{0.02}} &
  0.05 &
  0.29 &
  \multicolumn{1}{l}{\textbf{0.09}} &
  \textbf{0.28} &
  \multicolumn{1}{l}{0.16} &
  \textbf{0.23} &
  \multicolumn{1}{l}{0.03} &
  0.28 &
  0.27 &
  0.14 &
  0.68 &
  0.66 &
  0.00 &
  0.00 &
  0.09 &
  0.23 &
  0.55 &
  0.53 &
  \textbf{0.24} &
  0.14 &
  0.46 &
  - &
  - &
  - &
  - \\
Narayanan \textit{et al.} \cite{narayanan2019dynamic} &
  \multicolumn{1}{l}{0.30} &
  0.19 &
  \multicolumn{1}{l}{0.24} &
  0.24 &
  \multicolumn{1}{l}{\textbf{0.02}} &
  \textbf{0.06} &
  \textbf{0.38} &
  \multicolumn{1}{l}{0.056} &
  0.08 &
  \multicolumn{1}{l}{0.13} &
  0.16 &
  \multicolumn{1}{l}{0.08} &
  0.16 &
  0.23 &
  \textbf{0.31} &
  0.70 &
  0.67 &
  0.00 &
  0.00 &
  0.00 &
  0.42 &
  0.58 &
  0.59 &
  0.23 &
  \textbf{0.47} &
  0.46 &
  - &
  - &
  - &
  - \\
Single-Task &
  \multicolumn{1}{l}{0.25} &
  0.17 &
  \multicolumn{1}{l}{0.21} &
  0.18 &
  \multicolumn{1}{l}{0.00} &
  0.04 &
  0.37 &
  \multicolumn{1}{l}{0.04} &
  0.06 &
  \multicolumn{1}{l}{0.06} &
  0.16 &
  \multicolumn{1}{l}{0.02} &
  0.24 &
  0.24 &
  0.23 &
  0.71 &
  0.62 &
  0.00 &
  0.00 &
  0.00 &
  0.17 &
  0.40 &
  0.48 &
  0.00 &
  0.22 &
  0.00 &
  0.00 &
  \textbf{0.5} &
  \textbf{0.57} &
  0.89 \\
Multi-Task &
  \multicolumn{1}{l}{0.28} &
  0.17 &
  \multicolumn{1}{l}{0.23} &
  0.21 &
  \multicolumn{1}{l}{0.00} &
  0.02 &
  0.36 &
  \multicolumn{1}{l}{0.03} &
  0.03 &
  \multicolumn{1}{l}{0.07} &
  0.15 &
  \multicolumn{1}{l}{0.00} &
  0.23 &
  0.20 &
  0.20 &
  0.61 &
  0.68 &
  0.00 &
  0.00 &
  0.00 &
  0.28 &
  0.48 &
  0.54 &
  0.00 &
  0.11 &
  0.00 &
  0.00 &
  0.00 &
  0.00 &
  0.83 \\
Ours &
  \multicolumn{1}{l}{0.33} &
  0.19 &
  \multicolumn{1}{l}{\textbf{0.34}} &
  0.27 &
  \multicolumn{1}{l}{0.00} &
  0.04 &
  \textbf{0.38} &
  \multicolumn{1}{l}{0.04} &
  0.06 &
  \multicolumn{1}{l}{\textbf{0.26}} &
  0.18 &
  \multicolumn{1}{l}{\textbf{0.11}} &
  \textbf{0.38} &
  \textbf{0.28} &
  0.14 &
  \textbf{0.78} &
  \textbf{0.79} &
  0.00 &
  \textbf{0.06} &
  0.00 &
  \textbf{0.47} &
  \textbf{0.59} &
  \textbf{0.60} &
  0.10 &
  0.35 &
  \textbf{0.52} &
  0.00 &
  \textbf{0.5} &
  \textbf{0.57} &
  \textbf{0.90} \\ \hline
 &
  \multicolumn{9}{c|}{\textbf{Weather}} &
  \multicolumn{5}{c|}{\textbf{Surface}} &
  \multicolumn{16}{c|}{\textbf{Environment}} \\ \cline{2-31} 
\multicolumn{1}{c|}{} &
  \multicolumn{2}{c}{Sunny} &
  \multicolumn{2}{c}{Foggy} &
  \multicolumn{3}{c}{Cloudy} &
  \multicolumn{2}{c|}{Rainy} &
  \multicolumn{2}{c}{Dry} &
  \multicolumn{3}{c|}{Wet} &
  \multicolumn{3}{c}{Local} &
  \multicolumn{3}{c}{Highway} &
  \multicolumn{3}{c}{Ramp} &
  \multicolumn{3}{c}{Rural} &
  \multicolumn{4}{c|}{Urban} \\ \hline
Bi-LSTM \cite{yue2015beyond} &
  \multicolumn{2}{l}{-} &
  \multicolumn{2}{l}{-} &
  \multicolumn{3}{l}{-} &
  \multicolumn{2}{l|}{-} &
  \multicolumn{2}{l}{-} &
  \multicolumn{3}{l|}{-} &
  \multicolumn{3}{l}{-} &
  \multicolumn{3}{l}{-} &
  \multicolumn{3}{l}{-} &
  \multicolumn{3}{l}{-} &
  \multicolumn{4}{l|}{-} \\
Narayanan \textit{et al.} \cite{narayanan2019dynamic} &
  \multicolumn{2}{c}{0.86} &
  \multicolumn{2}{c}{\textbf{0.83}} &
  \multicolumn{3}{c}{-} &
  \multicolumn{2}{c|}{0.83} &
  \multicolumn{2}{c}{0.93} &
  \multicolumn{3}{c|}{\textbf{0.92}} &
  \multicolumn{3}{c}{0.33} &
  \multicolumn{3}{c}{0.91} &
  \multicolumn{3}{c}{0.20} &
  \multicolumn{3}{c}{-} &
  \multicolumn{4}{c|}{\textbf{0.83}} \\
Single-Task &
  \multicolumn{2}{c}{0.86} &
  \multicolumn{2}{c}{\textbf{0.83}} &
  \multicolumn{3}{c}{0.81} &
  \multicolumn{2}{c|}{0.83} &
  \multicolumn{2}{c}{0.93} &
  \multicolumn{3}{c|}{\textbf{0.92}} &
  \multicolumn{3}{c}{0.33} &
  \multicolumn{3}{c}{0.91} &
  \multicolumn{3}{c}{0.20} &
  \multicolumn{3}{c}{0.08} &
  \multicolumn{4}{c|}{\textbf{0.83}} \\
Multi-Task &
  \multicolumn{2}{c}{0.88} &
  \multicolumn{2}{c}{0.25} &
  \multicolumn{3}{c}{0.79} &
  \multicolumn{2}{c|}{0.86} &
  \multicolumn{2}{c}{0.93} &
  \multicolumn{3}{c|}{0.85} &
  \multicolumn{3}{c}{0.32} &
  \multicolumn{3}{c}{0.92} &
  \multicolumn{3}{c}{0.21} &
  \multicolumn{3}{c}{0.07} &
  \multicolumn{4}{c|}{0.80} \\
Ours &
  \multicolumn{2}{c}{\textbf{0.9320}} &
  \multicolumn{2}{c}{0.4495} &
  \multicolumn{3}{c}{\textbf{0.8398}} &
  \multicolumn{2}{c|}{\textbf{0.9699}} &
  \multicolumn{2}{c}{\textbf{0.9603}} &
  \multicolumn{3}{c|}{0.8617} &
  \multicolumn{3}{c}{\textbf{0.36}} &
  \multicolumn{3}{c}{\textbf{0.92}} &
  \multicolumn{3}{c}{\textbf{0.21}} &
  \multicolumn{3}{c}{\textbf{0.16}} &
  \multicolumn{4}{c|}{0.81} \\ \hline
\end{tabular}%
}
\end{table*}

\begin{table}[]
\vspace{-0.2cm}
\caption{Traffic scene recognition category-wise F-Scores comparison of method with existing methods on HSD. For fair comparisons, the \textit{Tunnel}, \textit{Zebra Crossing}, \textit{Cloudy}, and \textit{Rural} classes, which were not used in previous studies, were excluded from the calculations.}
\label{tab:my-table}
\begin{tabular}{l|llll}
               & Road Place     & Weather       & Surface  & Enviroment \\ \hline
Bi-LSTM \cite{yue2015beyond}        & 0.275          & -             & -             & -               \\
Narayanan \textit{et al.} \cite{narayanan2019dynamic}           & 0.285          & 0.91          & 0.95          & 0.56            \\
ResNet (STL)   & 0.263          & \textbf{0.93} & \textbf{0.96} & 0.73            \\
DenseNet (STL) & 0.229          & 0.88          & 0.86          & 0.68            \\
Baseline (MTL) & 0.252          & 0.91          & 0.93          & 0.64            \\ \hline
Ours           & \textbf{0.303} & \textbf{0.92} & \textbf{0.95} & \textbf{0.76}  
\end{tabular}
\end{table}

\subsection{Comparison with Other Methods} 
    We demonstrate the effectiveness of our MI-based MTL approach for the application of traffic scene recognition \cite{narayanan2019dynamic}. Table 2 shows the task-wise F-Scores and the recognition performance generally outperforms the existing methods across most of tasks. This mainly due to the fact that traffic scene samples are processed with context-aware feature representations. Note that what we want to highlight in HSD benchmark (especially notice the difference between Single-Task \cite{narayanan2019dynamic} and ours) is that the proposed method can benefit from MI based MTL framework which captures the context-aware properties of images despite only one optimization formula and less memory volume.

\section{Discussion}
\vspace{-0.1cm}
\subsection{Visualizing Analysis}
    Figure 5,6, and 7 indicate out method is significantly better than existing MTL method. Our visualization results based on t-SNE \cite{maaten2008visualizing} indicate that our approach proves the ability of discriminative representation and the shared encoder can have a better context-awareness ability. 
    
\subsection{Limitations}
    \vspace{-0.1cm}
    Here, we discuss two challenging problems, for which our method fails to recognize prediction. First, our method may fail to worst performance about classes with an extremely small number of data samples, \textit{e.g.,} Intersection-5 and Merge Left in the road place and ramp and rural in the road environment. In fact, the performance imbalance caused by this data imbalance can be found not only in our method but also in all existing methods. Second, our approach may wrongly recognize temporal classes in road place category. To address this issue, we believe that an online action proposal technique is needed for the MTL network to learn temporal-aware representations. Besides, we may carefully examine other possible mechanisms to avoid manual annotations for collecting more data.

\section{CONCLUSIONS}
In this paper, we have introduced a novel context-aware multi-task network for traffic scene recognition in autonomous vehicles. By calculating an information-theoretically motivated mutual information constraint, it is able to make use of task-invariant commonality and task-specific uniqueness properties, both of which are fundamental to visual recognition.  Moreover, we introduce a multi-objective loss to the multi-task network, which can optimize an end-to-end trainable multi-task network without additional task-specific network. Extensive experiments on a large-scale dataset HSD demonstrate our method can convincingly improve the performance over baseline methods. Furthermore, our proposed framework significantly performs favorably against the state-of-the-art methods in both single-task and multi-task settings. For future directions, we potentially expect this approach to be integrated into ADAS applications on farther ranges and various categories.

\addtolength{\textheight}{-12cm}   




\section*{ACKNOWLEDGMENT}

This work was partly supported by Institute of Information \& Communications Technology Planning \& Evaluation (IITP) grant funded by the Korea government (MSIT) (No.2014-3-00077, AI National Strategy Project) and the National Research Foundation of Korea (NRF) grant funded by the Korea government (MSIT) (No. 2019R1A2C2087489).


\addtolength{\textheight}{12cm}
\bibliography{egbib}
\bibliographystyle{ieeetran}

\end{document}